\documentclass[journal]{IEEEtran}
\usepackage{amsmath,amsfonts}
\usepackage{algorithmic}
\usepackage{algorithm}
\usepackage{array}
\usepackage[caption=false,font=normalsize,labelfont=sf,textfont=sf]{subfig}
\usepackage{textcomp}
\usepackage{stfloats}
\usepackage{url}
\usepackage{verbatim}
\usepackage{graphicx}
\usepackage{cite}
\usepackage{orcidlink}
\usepackage{tikz}
\def\BibTeX{{\rm B\kern-.05em{\sc i\kern-.025em b}\kern-.08em
    T\kern-.1667em\lower.7ex\hbox{E}\kern-.125emX}}
\usepackage{balance}
\hypersetup{
colorlinks=true,
linkcolor=black,
citecolor=black
}
\begin{document}
\title{Adaptive Traffic Element-Based Streetlight Control Using Neighbor Discovery Algorithm Based on IoT Events}
\author{
Yupeng Tan\hspace{-1.5mm}$^{~\orcidlink{0009-0002-5888-3945}}$, 
Sheng Xu\hspace{-1.5mm}$^{~\orcidlink{0000-0002-6314-8104}}$, and 
Chengyue Su
\thanks{This work has been submitted to the IEEE for possible publication. Copyright may be transferred without notice, after which this version may no longer be accessible.
}}

\markboth{IEEE INTERNET OF THINGS JOURNAL,~Vol.~18, No.~9, September~2020}%
{How to Use the IEEEtran \LaTeX \ Templates}

\maketitle

\begin{abstract}
Intelligent streetlight systems divide the streetlight network into multiple sectors, activating only the streetlights in the corresponding sectors when traffic elements pass by, rather than all streetlights, effectively reducing energy waste. This strategy requires streetlights to understand their neighbor relationships to illuminate only the streetlights in their respective sectors. However, manually configuring the neighbor relationships for a large number of streetlights in complex large-scale road streetlight networks is cumbersome and prone to errors. Due to the crisscrossing nature of roads, it is also difficult to determine the neighbor relationships using GPS or communication positioning. In response to these issues, this article proposes a systematic approach to model the streetlight network as a social network and construct a neighbor relationship probabilistic graph using IoT event records of streetlights detecting traffic elements. Based on this, a multi-objective genetic algorithm based probabilistic graph clustering method is designed to discover the neighbor relationships of streetlights. Considering the characteristic that pedestrians and vehicles usually move at a constant speed on a section of a road, speed consistency is introduced as an optimization objective, which, together with traditional similarity measures, forms a multi-objective function, enhancing the accuracy of neighbor relationship discovery. Extensive experiments on simulation datasets were conducted, comparing the proposed algorithm with other probabilistic graph clustering algorithms. The results demonstrate that the proposed algorithm can more accurately identify the neighbor relationships of streetlights compared to other algorithms, effectively achieving adaptive streetlight control for traffic elements.
\end{abstract}


\section{Introduction}
\IEEEPARstart{B}{y} intelligently adjusting the brightness levels of LEDs, intelligent streetlight systems can minimize lighting power consumption while ensuring the quality of illumination \cite{1,2}. However, on public roads, lighting demands are dynamically changing due to the random movement of traffic elements such as pedestrians and vehicles. Currently, intelligent streetlight systems based on Internet of Things (IoT) technology are widely applied, enabling the construction of distributed wireless sensor networks that can not only monitor streetlight status but also detect traffic flow and environmental conditions \cite{3,4}. Adaptive lighting for traffic elements is an effective method for energy-saving lighting, with the fundamental concept being that high brightness of streetlights is not required when traffic elements are not in the vicinity. Therefore, a mechanism has been proposed to divide the streetlight network into multiple sectors, activating streetlights in specific sectors only when traffic elements pass by, rather than all streetlights \cite{5}. This sectorized streetlight system can automatically reduce the illumination brightness of sector streetlights after traffic elements leave the sector, meeting both energy-saving and safety needs.

Intelligent streetlight systems adjust illumination brightness through message relay over wireless networks. When a node detects traffic elements, it sends a message into the network, and neighbor nodes that receive the message relay it, thereby completing the adaptive seamless lighting control for traffic elements \cite{6}. A sector-based system, upon detecting traffic elements, adjusts the illumination brightness of multiple sectors in the direction of the traffic elements' movement, providing a seamless lighting experience for drivers and reducing discomfort caused by frequent on-off switching of streetlights \cite{5} However, for complex road networks with crisscrossing paths, traditional relay strategies may lead to imprecision in information transmission, as advertising affects all nodes within the communication range, not just those located on the road where the traffic elements are present, as shown in fig. 1. To achieve accurate lighting control, streetlight nodes must accurately understand their neighbor relationships, including whether they are on the same road or within the same sector. In large-scale streetlight networks, relying on GPS or communication positioning technologies to determine the neighbor relationships of nodes can be unreliable due to the crisscrossing nature of roads, and manually configuring neighbor relationships can be both cumbersome and prone to errors, limiting the system's scalability and reliability.

\begin{figure}[!t]
\centering
\includegraphics[width=2.5in]{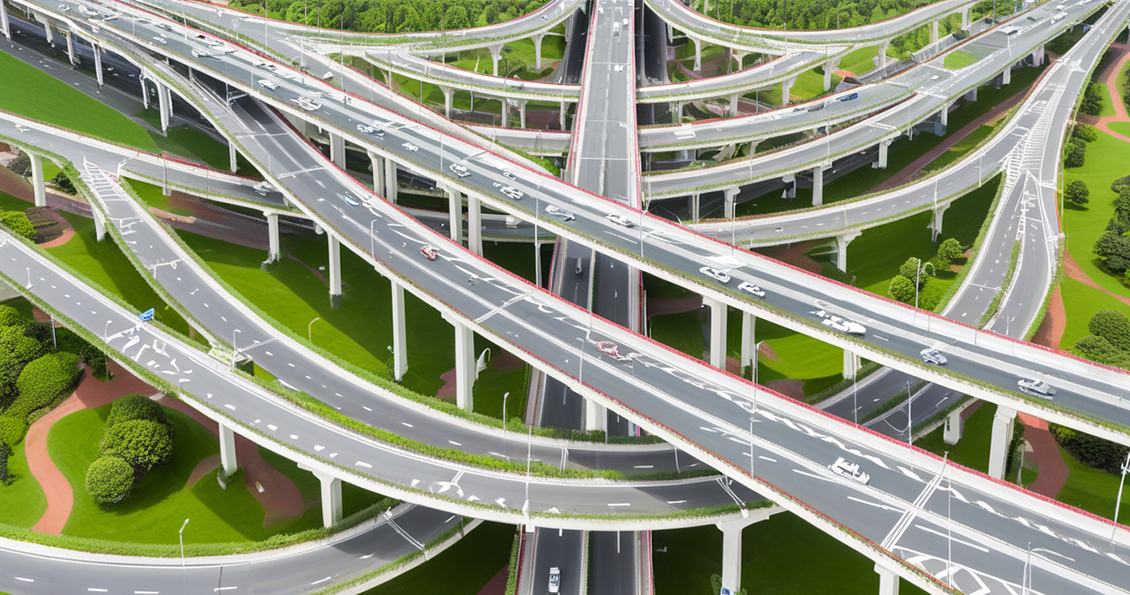}
\caption{Intersecting road network.}
\label{fig1}
\end{figure}

Addressing the aforementioned issues, this article proposes a systematic approach that models the streetlight network as a social network and employs clustering algorithms to mine the neighbor relationships of streetlights, thereby dividing the sectors where nodes are located and achieving accurate control of the streetlight network. In social networks, a community is a group of nodes that exhibit a high degree of interconnectivity among themselves and a lower level of interaction with nodes outside the group \cite{7}.Graph models, as a tool for modeling, are widely used to depict and analyze large data networks. However, the uncertainty of graph data is an issue that cannot be ignored, which may stem from measurement noise, inaccurate information sources, and other factors \cite{8,9} . Probabilistic graph models provide a means of modeling natural phenomena with uncertain interactions by assigning existence probabilities to each edge. In the field of probabilistic graph data mining, clustering is a common technique whose main goal is to gather similar nodes into clusters and assign dissimilar nodes to different clusters \cite{10}. The designed streetlight system consists of multiple streetlight nodes equipped with millimeter-wave radar sensors and Bluetooth Low Energy (BLE) modules. The millimeter-wave radar sensors are used to detect nearby pedestrians and vehicles, among other traffic elements. When traffic elements approach a node, the node records the event and advertises a message. As traffic elements move, subsequent nodes also record the event and continue advertising when they detect traffic elements. The series of advertising generate multiple records that are periodically uploaded to the gateway for mining the neighbor relationships between streetlight nodes. Considering the uneven distribution of nodes, unstable communication, and the existence of advertising storms, the relationships between nodes are uncertain; hence, this article uses probabilistic graph to model the streetlight network.

To mine the neighbor relationships between nodes, this article proposes a probabilistic graph clustering algorithm based on Multi-Objective Genetic Algorithm (MOGA) \cite{11}. Building upon the work in \cite{12}, the proposed algorithm employs a multi-population based GA \cite{13} for enhancing the accuracy of clustering results. In the streetlight network, when traffic elements pass by two streetlight nodes in succession, the latter node can deduce the movement time of the traffic elements, i.e., the time interval between two advertising, based on the advertising time of the former node and the detection time of its own. Observations indicate that vehicles typically move at a constant speed on a section of road until they encounter special locations such as junctions or corners. Additionally, to ensure uniformity of road illumination, streetlights are often evenly distributed along the road. This implies that the movement time of traffic elements between adjacent streetlight nodes on the same section of a road should exhibit a high degree of consistency. Based on this characteristic, this article introduces speed consistency as a secondary optimization objective, which, together with traditional similarity measures, forms a multi-objective optimization function to guide the search process of the GA.

In summary, the main contributions of this article are as follows:

\begin{enumerate}
\item{A modeling method is proposed that does not rely on the geographical location information of streetlight nodes but instead models the streetlight network as a social network represented by a probabilistic graph based on IoT event records. By applying clustering methods, it is able to automatically discover the neighbor relationships between streetlight nodes, thereby achieving adaptive lighting control for traffic elements.}
\item{A speed consistency measure is introduced as a secondary optimization objective, and a probabilistic graph clustering algorithm based on MOGA is constructed. This innovation enhances the accuracy of the clustering algorithm, allowing it to more accurately identify the neighbor relationships between streetlight nodes.}
\item{Simulation datasets were generated based on various scenarios, and extensive experiments were conducted on the proposed clustering algorithm. Comparisons were made with other probabilistic graph clustering algorithms, validating the effectiveness and superiority of the proposed algorithm.}
\end{enumerate}

The remainder of this article is organized as follows: Section 2 provides an overview of literature related to this article, Section 3 introduces the definitions related to probabilistic graphs. Section 4 details the proposed methods, including the modeling approach for social networks and the components of the GA. Section 5 presents the detailed experimental results, while Section 6 summarizes the work of this article.

\section{Related work}
\subsection{Adaptive Lighting for Traffic Elements}
In recent years, with the maturation of information and communication technologies, an increasing number of studies have integrated wireless modules into streetlight systems. This integration has not only achieved more precise and adaptive lighting management but also further reduced energy consumption while ensuring safety \cite{14,15}. The operation of street lights should be automatically adjusted based on moving objects to further reduce energy consumption \cite{16}. To this end, various sensing mechanisms have been proposed that can detect traffic elements and automatically adjust lighting levels accordingly. Among them, the mobile lighting strategy provides necessary illumination only around traffic elements by tracking them \cite{17}. This mobile object-centered control method minimizes energy consumption by monitoring the dynamics of traffic elements.

Many intelligent streetlight systems adjust the brightness of lighting through message relay in wireless networks. The scheme in \cite{18} intelligently regulates lighting according to the different needs of pedestrians and vehicles, ensuring that pedestrians receive adequate illumination within a 150-meter range, while providing comprehensive lighting coverage of 100 meters ahead and behind for vehicles. \cite{17} does not differentiate between pedestrians, vehicles, or other traffic elements, but determines the range of lighting activation solely based on their speed and direction. \cite{19} focuses on indoor lighting control, dynamically adjusting the lighting range based on the position of pedestrians to cover most of their field of view. \cite{20} detects the type of moving objects through sensor modules and transmits the detection results via the network to control the number of street lights activated.

In addition to message relay-based schemes, some studies have introduced sectorization methods. In the system proposed in \cite{21}, when two nodes within an area consecutively detect pedestrians, the lighting level of all nodes increases, even if some nodes have not yet detected pedestrians. \cite{22} places multiple streetlights in the junction area to form a street light island. When the sensors at the intersection detect traffic elements, the control unit turns on all streetlights and keeps them on for a fixed period. \cite{5} divides the entire highway into multiple sectors and turns on all street lights in the sector where the anchor node is located, as well as some street lights in the next sector, when a vehicle approaches the anchor node. \cite{6} adopts a sector-based control during the night, lighting up half of the street lights at regular intervals, while the other half is partially turned on in advance based on the location of traffic elements. As mentioned earlier, sectorization methods require determining the neighbor relationships of nodes, which is also the starting point of this article.

\subsection{Probabilistic Graph Clustering}
Cluster analysis, as one of the core tasks in the field of data mining and machine learning, has been deeply studied for half a century. Despite this, studies focusing on clustering problems of uncertain graph remain sparse, yet the domain is witnessing swift growth. Some algorithms have extended the original deterministic graph clustering techniques. For instance, \cite{23} proposed the pKwikCluster algorithm, which randomly chooses a node each iteration and groups this node and all its neighboring nodes with a probability greater than 0.5 into a cluster, until all nodes have been traversed. \cite{24} proposed the USCAN algorithm, which introduces the concept of reliable structural similarity and efficiently calculates the similarity probability between nodes through dynamic programming algorithms. \cite{25} improved the traditional k-median and k-center algorithms by providing approximate solutions through graph sampling and greedy strategies, without relying on connectivity predictors. \cite{26} proposed a density-based clustering algorithm, DBCLPG, which utilizes the density of the graph and neighborhood information of nodes for clustering.

In the field of probabilistic graph clustering, the works in \cite{27}, \cite{12} and \cite{28} are closely related to this article. \cite{27} proposed a clustering method for large-scale probabilistic graphs based on a multi-population GA, known as the EA-CPG (Evolutionary Algorithm based Clustering of Probabilistic Graphs). This algorithm transforms the probabilistic graph into multiple deterministic graphs through different thresholds and generates multiple initial populations based on this, where each chromosome acts as a complete clustering solution. During the genetic iteration process, the EA-CPG employs the pKwikCluster algorithm to adjust the clustering results. Building upon the EA-CPG, \cite{12} proposed an improved algorithm called EEA-CPG (Extended EA-CPG). The improvements of this algorithm include: (1) employing a hybrid initialization strategy to generate initial populations; (2) utilizing a higher order random walk model to assess the similarity between nodes; (3) introducing a local search process based on the similarity between nodes. More recently, \cite{28} proposed a clustering method based on the Ant Colony Optimization (ACO) algorithm, referred to as the ACO-CUG. This algorithm adopts a method based on the Farthest Distance Cluster Center (FDCC) for initializing the pheromone matrix, providing a good initial solution.

Addressing the complex large-scale streetlight networks, this article employs a probabilistic graph clustering algorithm to segment the nodes into different sectors, achieving intelligent sectorization of the streetlight network. The clustering algorithm based on node similarity and GA used in \cite{12} is conceptually similar to the proposed algorithm, but there are several key differences. This article innovates in the design of the objective function to evaluate the metric of speed consistency. Furthermore, the proposed method provides an automated clustering solution, eliminating the need for users to pre-specify the number of clusters, making it suitable for scenarios where the ideal number of clusters is unknown.

\section{Probabilistic graphs}
\textbf{Definition 1} (\textit{Probabilistic Graph}): A probabilistic graph $G^P$ is represented as $G^P=(V,E,P)$, where $V$ denotes the set of nodes, $E$ denotes the set of edges, and $P$ denotes the set of probabilities associated with the edges. Each edge $e_{u,v} \in E$ in the graph is assigned a probability $p_{u,v} \in P$, indicating the likelihood of the existence of edge $e_{u,v}$ in graph $G^P$.

\textbf{Definition 2} (\textit{Deterministic Graph}): Given a probabilistic graph $G^P$, a deterministic graph $G^D=(V,E)$ is defined as a subset of $G^P$, where each edge $e_{u,v}\in E$ represents a binary relationship (deterministic connection) between nodes $u$ and $v$.

\textbf{Definition 3} (\textit{Weighted Deterministic Graph}): A weighted deterministic graph $G^W=(V,E,\lambda,W)$ is based on a deterministic graph, assigning a weight $w_{u,v}$ to each edge $e_{u,v}\in E$, which is not greater than $\lambda$, and $W$ represents the set of all weights. For any $G^W\subseteq G^P$, this article defines the weight $w_{u,v}$ of an edge as the probability $p_{u,v}$ of the existence of the edge.

\textbf{Definition 4} (\textit{Simple Path}): Given a graph $G^P=(V,E,P)$, a simple path (also known as an order-1 path) between nodes $v_i$ and $v_j$, $v_i,v_j\in V$, is a sequence of nodes represented as $v_i\overset{e_{v_i,v_j}}{\longrightarrow}v_j$. For the special case of $v_i=v_j$, it is referred to as a self-loop in this article.

\textbf{Definition 5} (\textit{Higher Order Path}): Given a graph $G^P = (V, E, P)$, a $t^{th}$ order path between nodes $v_i$ and $v_j$, $v_i, v_j \in V$, is represented as $path^t = (v_i, v_{i+1}, ..., v_{i+t-1}, v_j)$. $path^t$ is a sequence of nodes such that $\forall v_k \in path^t(v_i, v_j)$ and $\forall v_k \in path^{t-1}(v_i, v_{j-1})$, $\exists e_{k,k+1} \in E$.

To convert a probabilistic graph into a weighted deterministic one, the thresholds proposed in \cite{27} are utilized. Specifically, all edges in the probabilistic graph $G^P$ are applied to a preset threshold $k$. An edge is kept and assigned a probability as weight only when its probability is greater than or equal to $k$. Fig. 2 illustrates this process with a simple example. Fig. 2(a) presents a probabilistic graph comprising 6 nodes and 9 edges, with the probability of each edge indicated. Fig. 2(b), 2(c), and 2(d) demonstrate the weighted deterministic graphs derived from applying different thresholds (0.3, 0.5, and 0.8), respectively. It is observable from this example that a lower threshold leads to a greater number of edges, with a lower degree of certainty. In contrast, a higher threshold reduces the number of edges, increasing the certainty of those edges.

\begin{figure}[!t]
\centering
\subfloat[]{\includegraphics[width=1.5in]{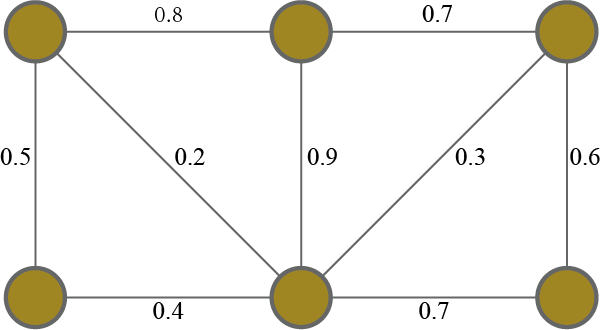}
\label{fig2a}}
\hfil
\subfloat[]{\includegraphics[width=1.5in]{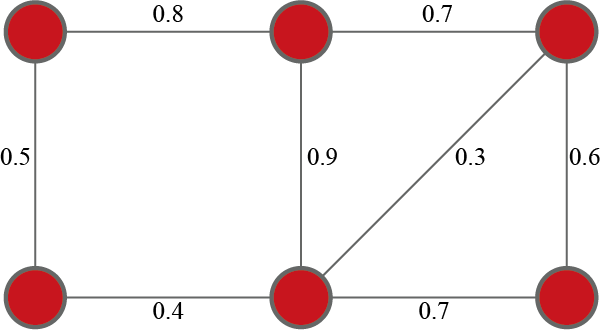}
\label{fig2b}}
\\
\subfloat[]{\includegraphics[width=1.5in]{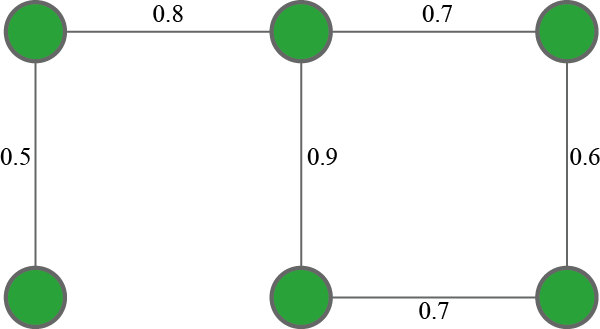}
\label{fig2c}}
\hfil
\subfloat[]{\includegraphics[width=1.5in]{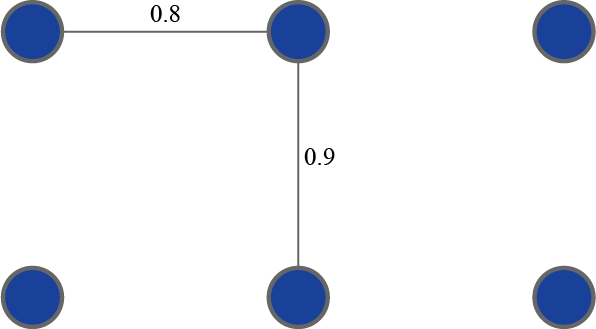}
\label{fig2d}}
\caption{Probabilistic graph and weighted deterministic graphs. (a) Probabilistic graph. (b) Deterministic graph with $\lambda=0.3$. (c) Deterministic graph with $\lambda=0.5$. (d) Deterministic graph with $\lambda=0.8$.}
\label{fig2}
\end{figure}

\section{Proposed method}
\subsection{Social Network Modeling Method}
The streetlight system designed in this article comprises multiple streetlight nodes, each equipped with a BLE module. Through these modules, the streetlight nodes are capable of communicating with each other, forming a BLE Mesh network. Each streetlight node is also integrated with a millimeter-wave radar sensor, which can sense traffic elements such as pedestrians and vehicles. In addition to the streetlight nodes, the system includes a gateway node responsible for collecting and processing data from the streetlight nodes.

The streetlight nodes adhere to a specific workflow. As shown in Fig. 3(a), when a streetlight node 1 detects the approach of a traffic element, it records the current time and advertises a message through the BLE Mesh network. The Time to Live (TTL) of the message is set to 1, ensuring that it is only received by nearby nodes. After node 2 receives the message, it temporarily stores the sender's MAC address, $MAC_1$, and the reception time, $t_1$, locally. As the traffic element continues to move and approaches node 2, this node stores the current time, $t_2$, and advertises a message, as depicted in Fig. 3(b). In addition to this, node 2 will correlate the two sets of data to generate an "association record" in the format $\{MAC_2,t_1,MAC_2,t_1\}$. This record indicates that nodes 2 and 1 may be in a neighbor relationship. As the traffic element continues to move, the streetlight nodes will generate multiple association records based on the aforementioned operations. These records collectively form a relation network that maps out the potential neighbor relationships between the nodes.

\begin{figure*}[!t]
\centering
\subfloat[]{\includegraphics[width=3.0in]{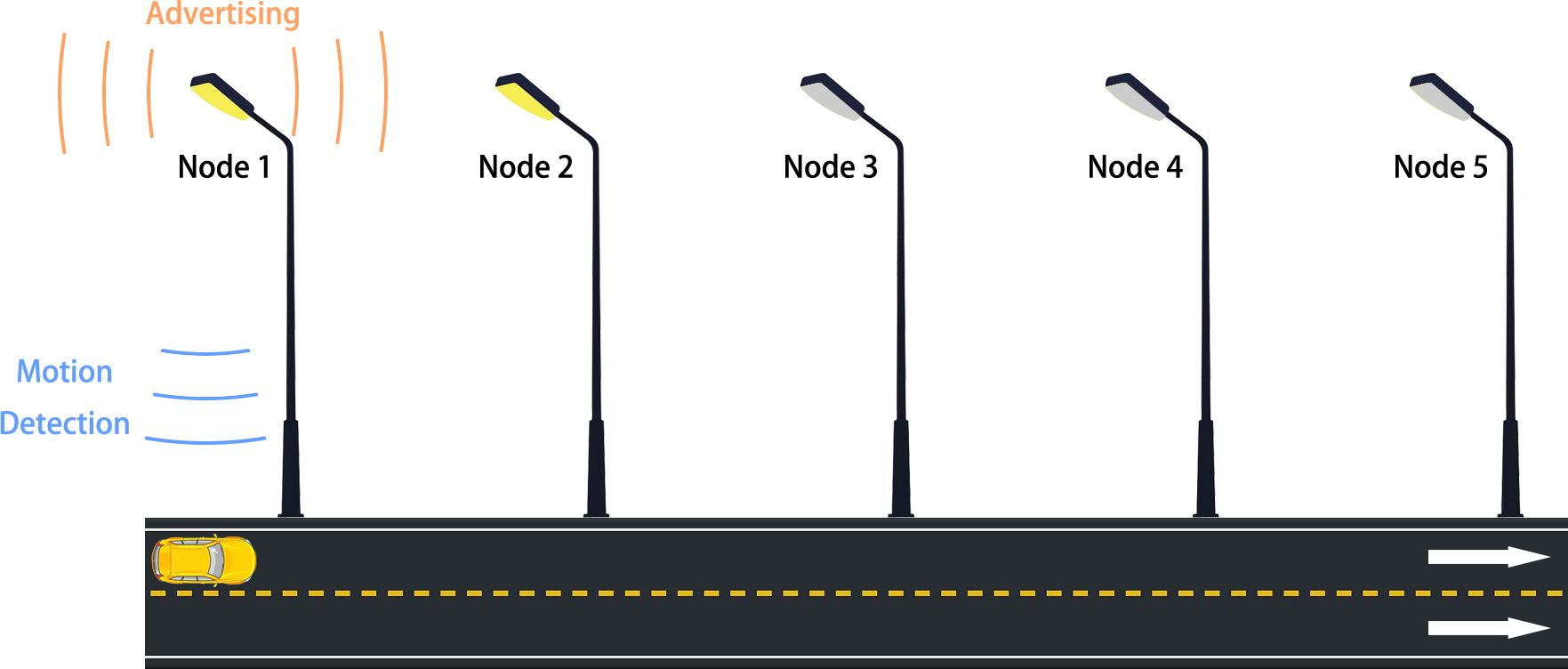}
\label{fig3a}}
\hfil
\subfloat[]{\includegraphics[width=3.0in]{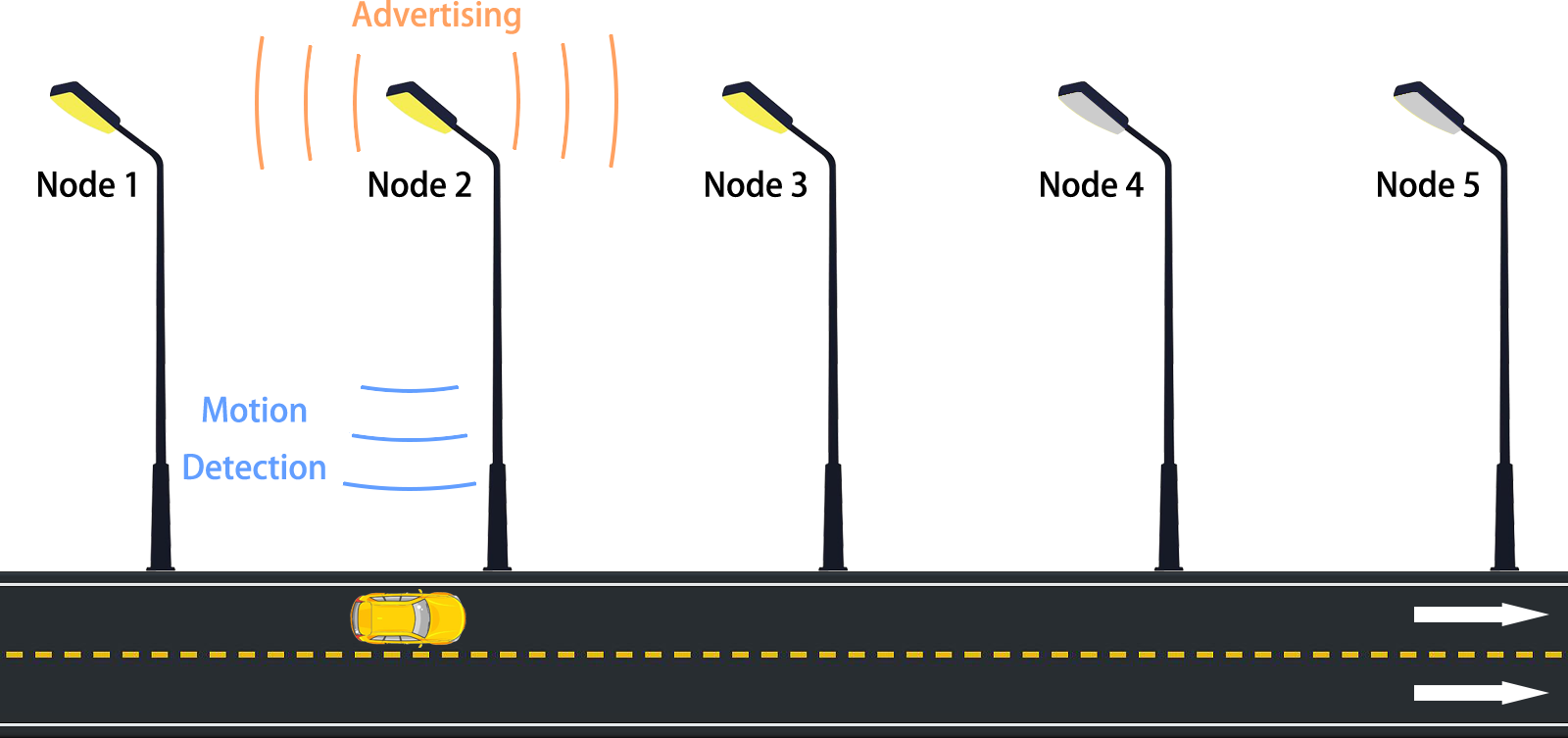}
\label{fig3b}}
\caption{Node detects a traffic element. (a) Node 1 detects a traffic element. (b) Node 2 detects a traffic element}
\label{fig3}
\end{figure*}

The streetlight system features a periodic data reporting mechanism. Every $T_{c}$ interval, the gateway requests all streetlights to sequentially report all association records generated within the past $T_{c}$ time period. These records are then centrally processed to construct a count matrix $C$. In matrix $C$, the element $c_{ij}$ represents the number of times node $j$ appears as the preceding node in the association records of node $i$. For instance, if the records of node $i$ contain $n$ association records of the form $\{MAC_{i}, t_{i}, MAC_{j}, t_{j}\}$, then $c_{ij} = n$. By normalizing the matrix $C$, each element $c_{ij}$ is transformed into a probability value $p_{ij}$. This transformation process takes into account the maximum value of the elements in each row, ensuring the rationality and comparability of the probability values. Through this processing, the count matrix $C$ is converted into the adjacency matrix $P$. Based on the adjacency matrix $P$, the gateway can construct a complete probabilistic graph.

In addition to the adjacency matrix $P$, the gateway also generates a time matrix $T$ to store the time differences in the association records, that is, the time intervals between two advertisings. The time matrix $T$ is a special matrix proposed for the scenario of this article and plays a key role in the calculation of the objective function. For $n$ identical association records $\{MAC_i, t_i^k, MAC_j, t_j^k\}, k=1,2,\ldots, n$, the time difference $t_{ij}^k = t_i^k - t_j^k$ is calculated for each record. Then, the average value $\overline{t_{ij}} = (\sum_{k=1}^{n} t_{ij}^k) / n$ is taken as the average advertising time interval between nodes $i$ and $j$, that is, $t_{ij} = \overline{t_{ij}}$. For ease of description, $t_{ij}$ will hereafter be referred to as the time interval between nodes $i$ and $j$. After processing all association records in the above manner, the gateway obtains the time intervals between any two nodes.

\subsection{Clustering Algorithm Based on MOGA}
The proposed method utilizes a clustering algorithm based on MOGA, which optimizes the initial population to yield the final clustering outcomes. The initialization of the population employs a rapid clustering mechanism coupled with a random strategy. Following the formation of the initial population, genetic operators are leveraged to iteratively refine the solution. In contrast to prior work \cite{12}, the innovation of the proposed algorithm is the incorporation of a novel objective function, which establishes a framework for multi-objective optimization. Moreover, the population initialization method has been modified, eliminating the need for users to pre-specify the number of clusters, thus making it suitable for scenarios where the number of clusters is not known a priori. The subsequent sections will provide a detailed exposition of the algorithm's operational principles, encompassing the construction of the mathematical model and the specific details of implementation.

\subsubsection{Proposed objective function}
Considering the characteristic that traffic elements generally move at a constant speed on a section of a road, the Speed Consistency Evaluation (SCE) is proposed as an objective function. This function assesses the consistency of time intervals between nodes within a cluster, specifically given by the following formula:

\begin{equation}
\label{eq1}
SCE(x_i) = \frac{\sum_{j=1}^{m} SC(C_j)}{m}
\end{equation}

\begin{equation}
\label{eq2}
SC(C_j) = 1 - \tanh\frac{\sigma(C_j)}{\mu(C_j)}
\end{equation}

\begin{equation}
\label{eq3}
\mu(C_j) = \frac{\sum_{l,k \in C_j} t_{lk}}{n_{C_j}}
\end{equation}

\begin{equation}
\label{eq4}
\sigma(C_j) = (\frac{\sum_{l, k \in C_j} (t_{lk} - \mu)^2} {n_{C_j}})^{\frac{1}{2}}
\end{equation}

where $x_i$ is the chromosome, $t_{lk}$ is the time interval between nodes $l$ and $k$, $n_{C_j}$ is the number of nodes in cluster $C_j$, $m$ is the number of clusters, and $\sigma(C_{j}) / \mu(C_j)$ is the coefficient of variation based on time intervals for cluster $C_j$.

For a cluster composed of nodes from the same section of a road, ideally, the time intervals between adjacent nodes should be consistent. In such cases, calculating the $SC$ based on any arbitrary selection of $t_lk$ could lead to anomalous scores and an inaccurate assessment of the quality of clustering results. Since nodes in probabilistic graph do not contain positional information, directly determining the neighbor relationships for all nodes is challenging. To address this issue, this article transforms the problem of determining the neighbor relationships within a cluster into the Traveling Salesman Problem (TSP) and solves it using the Christofides algorithm \cite{29}, thereby obtaining the optimal neighbor relationships among nodes within the cluster. The specific modeling and analysis are as follows.

For any cluster $C$, let $\rho$ be a permutation of all nodes in $C$, i.e., $\rho = (\rho_1, \rho_2, \ldots, \rho_{n_C})$, where $\rho_i$ represents the node in the $i^{\text{th}}$ position of the permutation. Let $T_C = \{t_{lk} \mid l, k \in C\}$ represent all time intervals within cluster $C$. For any valid permutation $\rho$, let $t_{\rho_i\rho_{i+1}}$ represent the time interval between adjacent nodes $\rho_i$ and $\rho_{i+1}$ in the permutation, and it satisfies $\forall t_{\rho_i\rho_{i+1}} \in T_C$. Assuming all nodes in cluster $C$ are located on the same road segment, the time intervals between all adjacent nodes will tend to be consistent, or close to a same value, defined as $t_C$. The time interval dissimilarity of permutation $\rho$ is defined as:

\begin{equation}
\label{eq5}
\text{DISS}(\rho) = \sum_{i=1}^{n_C-1} \left| t_{\rho_i\rho_{i+1}} - t_C \right|
\end{equation}

If there exists a permutation $\rho^*$ that minimizes the DISS, i.e.,

\begin{equation}
\label{eq6}
\rho^* = \underset{\rho}{\text{arg min}}(\text{DISS}(\rho))
\end{equation}

then the node neighbor relationships corresponding to $\rho^*$ are considered the best neighbor relationships for all nodes in cluster $C$. Given $t_C$, let $\Delta t_{\rho_i\rho_{i+1}} = |t_{\rho_i\rho_{i+1}} - t_C|$, equation (\ref{eq6}) can be rewritten as:

\begin{equation}
\label{eq7}
\rho^* = \underset{\rho}{\text{arg min}}\left(\sum_{i=1}^{n_C-1} \Delta t_{\rho_i\rho_{i+1}}\right)
\end{equation}

At this point, the problem transforms into finding a permutation $\rho^*$ that minimizes the sum of $\Delta t_{\rho_i\rho_{i+1}}$. If $\Delta t_{\rho_i\rho_{i+1}}$ is regarded as the weight of edge $e_{\rho_i\rho_{i+1}}$, then the problem can be translated into the Traveling Salesman Problem (TSP) without cycle. Using the Christofides algorithm to solve this can yield an optimal path that includes all nodes, i.e., the best permutation $\rho^*$.

The key to finding the optimal permutation $\rho^*$ lies in determining $t_C$. For a cluster with an unknown permutation, each $t_{lk}$ could potentially be $t_{\rho_i\rho_{i+1}}$, or in other words, each $t_{lk}$ could be close to $t_C$. Therefore, this article considers each $t_{lk}$ in the cluster as $t_C$ in turn and seeks the best permutation based on this assumption. After obtaining the permutations corresponding to each $t_{lk}$, the permutation with the lowest DISS according to equation (\ref{eq5}) is selected and taken as the optimal permutation $\rho^*$. The pseudocode for determining the best permutation of nodes within a cluster is shown in Algorithm \ref{alg1}.

\begin{algorithm}[H]
\caption{Determining the optimal permutation of nodes in a cluster.}\label{alg:alg1}
\begin{algorithmic}
\STATE \textbf{Input:} Nodes in cluster $(V_C)$, Time intervals $(T_C)$
\STATE \textbf{Output:} Best permutation $(\rho^*)$
\STATE 
\STATE \textbf{for} $t_{lk}$ in $T_C$ \textbf{do}:
\STATE \hspace{0.5cm} $t_C \gets t_{lk}$
\STATE \hspace{0.5cm} $\Delta T_C \gets \left\{ \left( |t_{lk} - t_C| \right) \mid t_{lk} \in T_C \right\}$
\STATE \hspace{0.5cm} Construct a weighted graph $G_C$ using $V_C$ and $\Delta T_C$
\STATE \hspace{0.5cm} Solve TSP of $G_C$ using christofides algorithm
\STATE \hspace{0.5cm} Save the permutation in $BEST_\rho$
\STATE \textbf{end for}
\STATE \textbf{Select} the best permutation $\rho^*$ in $BEST_\rho$ using (\ref{eq5})
\STATE \textbf{return} $\rho^*$
\end{algorithmic}
\label{alg1}
\end{algorithm}

\subsubsection{Proposed algorithm}
The overall process of the proposed clustering algorithm, based on MOGA, is as follows. First, given a probabilistic graph, six thresholds are independently applied to the graph to generate six subgraphs. For each subgraph, the higher order random walk similarity between all nodes is calculated. Then, six chromosome populations are initialized using a rapid clustering algorithm and random methods, with each population corresponding to a subgraph. Subsequently, in each iteration, local search and genetic operators are applied to all individuals in the population until the termination conditions are met. After the genetic iteration is terminated, the global best chromosome is determined based on the fitness value, and it is taken as the clustering result. The proposed algorithm is described in the following sections.

\begin{enumerate}
\item{
Chromosome encoding: A chromosome is an abstract representation of solution to a specific problem. For a graph $G$ containing $|V|$ nodes, a chromosome is defined as a one-dimensional array whose length is equal to the number of nodes. Each element in the array corresponds to a node in the graph, and the value of the element indicates the cluster number to which that node belongs. Fig. 4 illustrates the   structure of a chromosome with 7 nodes and its corresponding graph.

\begin{figure}[!t]
\centering
\subfloat[]{\includegraphics[width=1.5in]{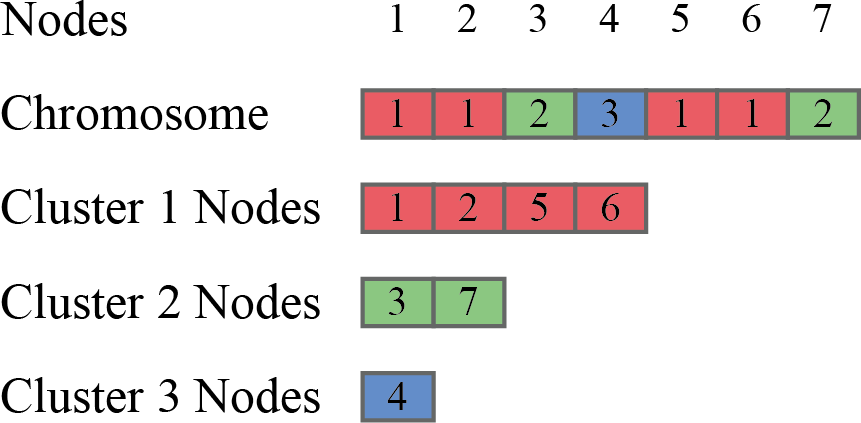}
\label{fig4a}}
\hfil
\subfloat[]{\includegraphics[width=1.5in]{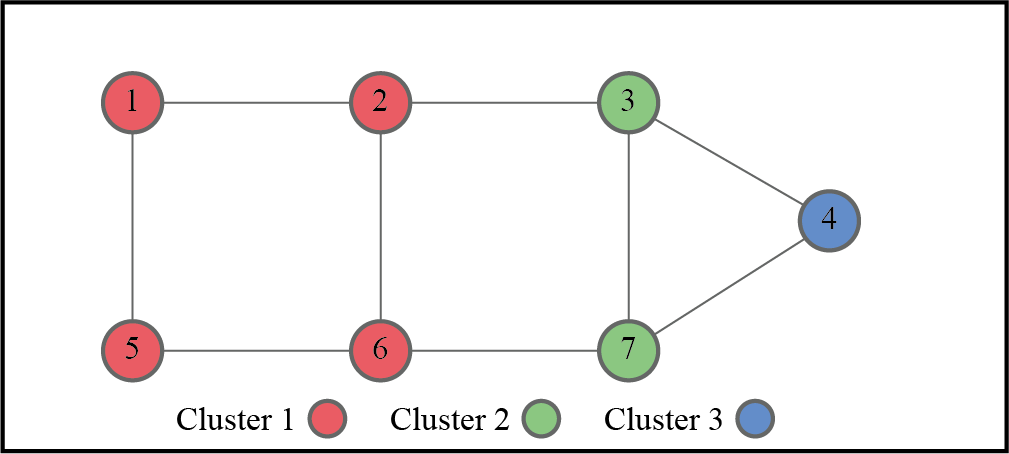}
\label{fig4b}}
\caption{The structure of a chromosome. (a) Vector form. (b) Graphical interpretation.}
\label{fig4}
\end{figure}
}
\item{
Graph initialization: In this phase, different deterministic versions $G$ of the input probabilistic graph $G^P$ are generated by applying a series of different thresholds $\lambda_i$. According to \cite{27}, six thresholds ranging from 0.3 to 0.8 with an interval of 0.1 were selected to generate six different subgraphs $G_i = (V, E, \lambda_i, W), i = 1, \ldots, 6$ of the probabilistic graph $G^P = (V, E, p)$.
}
\item{
Random walk: In this phase, the random walk (or random surfer) model proposed in \cite{12} is used to calculate the higher order random walk similarity between all nodes in each deterministic graph $G$. Specifically, the $t^{\text{th}}$ order similarity $\text{sim}^t(v_i, v_j)$ between nodes $v_i$ and $v_j$ is defined by the following formula:

\begin{equation}
\label{eq8}
\text{sim}^t(v_i, v_j) = \sum_{\alpha} \sum_{\gamma} \omega_{i\alpha} \cdot \text{sim}^{t-2}(v_{\alpha}, v_{\gamma}) \cdot \omega_{j\gamma}
\end{equation}

where $\omega_{ij}$ is defined as the probability of transitioning from node $v_i$ to $v_j$.
}
\item{
Population initialization: In this phase, for each deterministic graph $G$, 50\% of the population is initialized using a rapid clustering algorithm and random methods. The main difference from \cite{12} is that this article employs the pKwikCluster algorithm instead of the k-means algorithm to initialize 50\% of the population. The pKwikCluster algorithm is chosen because it does not require the number of clusters to be specified in advance, making it particularly suitable for scenarios where the number of clusters is unknown.
}
\item{
Local search: In this phase, a local search process is used to adjust the chromosomes in order to enhance the search depth of the algorithm. According to the method in \cite{12}, each node $x_{ij}$ in chromosome $x_i$ will be assigned a cluster label based on the following formula:

\begin{equation}
\label{eq9}
x_{ij} = \underset{\alpha}{\text{arg max}}\left(\sum_{v_l \in C_\alpha} \text{sim}^t(v_j, v_l)\right), \alpha = 1, \ldots, k
\end{equation}

where $C_\alpha$ is the group of nodes in cluster $\alpha$.
}
\item{
Objective function: The objective function is used to evaluate chromosomes in order to select the most promising individuals from the population. Drawing on the work in \cite{27} and \cite{12} , this article introduces two objective functions, the Dissimilarity Measure (DISIM) \cite{30} and the Distance Function (DIST) \cite{31}.

\begin{enumerate}
\item{
DISIM is designed to quantify the difference between the average dissimilarity among clusters and within clusters, and then normalize it. The calculation formula is as follows:

\begin{equation}
\label{eq10}
\text{DISIM}(x_i) = \frac{b(x_i) - a(x_i)}{\max\{a(x_i), b(x_i)\}}
\end{equation}

Where $b(x_i)$ represents the average dissimilarity of node $x_i$ relative to nodes in other clusters, and $a(x_i)$ represents the average dissimilarity of node $x_i$.
}
\item{
DIST is defined as the difference between the inter-cluster distance and the intra-cluster distance, with the calculation formula shown as follows:

\begin{equation}
\label{eq11}
\text{DIST}(x_i) = \sum_{j=1}^k \left[ D_{\text{inter}}(C_j) \cdot w - D_{\text{intra}}(C_j) \right]
\end{equation}

Where $x_i$ is the chromosome, $D_{\text{inter}}(C_j)$ is the inter-cluster distance of cluster $C_j$, $D_{\text{intra}}(C_j)$ is the intra-cluster distance of cluster $C_j$, $w$ is the weight parameter, and $k$ is the number of clusters. When the weight parameter $w$ takes a smaller value, the relative importance of $D_{\text{intra}}(C_j)$ is increased, which tends to produce a greater number of clusters, each of which is more compact, and vice versa.
}
\end{enumerate}
}
\item{
Multi-objective optimization: Due to the characteristic of time interval consistency being primarily targeted at a single section of a road, it is challenging for SCE to calculate the dissimilarity of time intervals between different clusters. When only considering the consistency within clusters, clusters with fewer nodes tend to have higher SCE scores, which is detrimental for longer sections of a road. DIST can adjust the compactness of clusters by modifying the value of the weight w, allowing it to flexibly adapt to roads of varying lengths. Therefore, this article employs both DIST and SCE to evaluate the quality of clustering.

To effectively combine the similarity measure and the speed consistency measure, both DIST and SCE must be optimized simultaneously. In multi-objective optimization problems, a common strategy is to assign a weight $\omega_i$ to each normalized objective function to transform the problem into an optimization problem with a single scalar objective function, as shown in the following equation:

\begin{equation}
\label{eq12}
\text{Fitness}(x_i) = \omega_1\text{DIST}(x_i) + \omega_2\text{SCE}(x_i)
\end{equation}

Unlike most literature works, in this article, the core function of DIST is to adjust the compactness of clusters, making it more in line with the characteristics of streetlight networks. That is, DIST and SCE do not exhibit the traditional mutually restrictive relationship but instead jointly guide the clustering process to achieve better clustering results. Furthermore, in practical application scenarios, prior information such as the geographical location and neighbor relationships of streetlight nodes is often unknown, which increases the difficulty of selecting the best solution from multiple Pareto optimal solutions. Based on the aforementioned concept, this article employs a fixed-weight method to guide the objective function in generating a single optimal solution. In this method, the weight $\omega_1$ is determined by experiments, while $\omega_2 = 1 - \omega_1$.
}
\item{
Selection: This article employs an elitist selection strategy, where the chromosomes are sorted by their fitness values, and the top 50\% of the performing individuals are selected as the parent population.
}
\item{
Crossover: This article utilizes a single-point crossover strategy. Specifically, for a randomly selected pair of parent chromosomes $x_i$ and $x_j$, a crossover point $C_p$ is randomly determined. The first offspring chromosome is constructed by combining the first half of $x_i$ with the second half of $x_j$, and the second offspring chromosome is constructed by combining the first half of $x_j$ with the second half of $x_i$.
}
\item{
Mutation: This article adopts a single-point mutation strategy. Specifically, for each offspring chromosome $x_i$, a random number $m_i$ between 0 and 100 is generated. If $m_i$ is lower than the preset mutation probability, a gene site on the chromosome is randomly selected and its value is changed to another valid value.
}
\end{enumerate}

The proposed algorithm essentially inherits from the EEA-CPG algorithm, but it has been improved in terms of the objective function and population initialization. In the proposed method, the pKwikCluster algorithm is used to initialize half of the population. The advantage of this algorithm is that it does not require the user to pre-specify the number of clusters, making it suitable for scenarios where the number of clusters is unknown. This article focuses on improving the objective function; the proposed SCE is based on the characteristic that traffic elements usually maintain a constant speed on a section of a road, measuring the quality of clustering results by assessing the consistency of time intervals within clusters. Compared to traditional objective functions, SCE is more in line with the characteristics of streetlight networks.

\section{Conclusion}
To address the cumbersome task of manually setting node neighbor relationships in streetlight networks, a method of modeling the streetlight network as a social network and representing it using a probabilistic graph model has been proposed in this article. This approach does not necessitate the geographical location information of streetlight nodes but dynamically adjusts the structure of the social network based on the actual flow of traffic elements, allowing it to adapt to various regions and traffic environments. Furthermore, a MOGA-based clustering algorithm has been proposeed for discovering community structures within the network. The key characteristic of this algorithm lies in the introduction of speed consistency as an optimization objective and the adjustment of the population initialization method. By examining the consistency of time intervals between nodes within clusters, the organization structure of clusters can be effectively evaluated and optimized. The proposed algorithm has been experimentally validated on multiple simulated datasets and compared with several existing probabilistic graph clustering methods. The experimental results demonstrate that, in the majority of datasets, the proposed algorithm can more accurately delineate sectors within the streetlight network.

In subsequent research, we plan to optimize the movement mechanism of virtual objects in the simulation. In the current mechanism, the speed of vehicles only changes at junctions and corners, while the speed of pedestrians remains constant, which may have limitations. Therefore, we will consider introducing complex factors such as traffic flow fluctuations and congestion to enhance the realism and accuracy of the dataset. Additionally, we plan to improve the G by adopting more advanced optimization techniques. The current GA mainly relies on basic genetic operators to adjust the population, which limits the breadth and depth of the search. To this end, we are considering introducing advanced techniques such as catastrophe and niche to enhance the search efficiency of the algorithm and maintain the diversity of the population.

\begin{IEEEbiographynophoto}{Yupeng Tan}
received the B.Sc. degree in optoelectronic information science and engineering from Guangdong University of Technology, Guangzhou, China, in 2018, where he is currently pursuing the M.E. degree with the School of Physics and Optoelectronic Engineering.

His research interests include intelligent lighting, Internet of Things, and clustering algorithms.
\end{IEEEbiographynophoto}

\begin{IEEEbiographynophoto}{Sheng Xu}
received the Ph.D. degree in Control Engineering from the South China University of Technology, Guangzhou, China, in 2008.

He is currently an assistant professor with the School of Physics and Optoelectronic Engineering, Guangdong University of Technology, Guangzhou, China. His research interests include machine vision and motion control.
\end{IEEEbiographynophoto}

\begin{IEEEbiographynophoto}{Chengyue Su}
received the Ph.D. degree in Astrophysics from Yunnan Observatories, Chinese Academy of Sciences, Kunming, china, in 2006.

He is currently a Professor and Master's Supervisor with the School of Advanced Manufacturing, Guangdong University of Technology, Guangzhou, China. He also served as as the Associate Dean of the School of Physics and Optoelectronic Engineering, and as a member of the University Laboratory Construction Committee and the Electrical and Electronic Experimental Teaching Steering Committee. He has been involved in two projects funded by the National Natural Science Foundation of China and provincial science and technology projects. His research interests include intelligent control, machine vision, and the Internet of Things.
\end{IEEEbiographynophoto}

\end{document}